\documentclass[10pt,a4paper]{article} 
\usepackage[margin=1.5cm]{geometry}
\usepackage[numbers]{natbib}
\usepackage{amsmath}
\usepackage{paralist}
\usepackage[hidelinks]{hyperref}
\hypersetup{colorlinks,linkcolor={blue},citecolor={blue},urlcolor={blue}}
\usepackage[utf8]{inputenc}
\usepackage{MnSymbol}
\usepackage{titlesec}

\titlespacing*{\section}{0pt}{2.5ex plus 1ex minus .2ex}{1.9ex plus .2ex}
\titlespacing*{\subsection}{0pt}{2.5ex plus 1ex minus .2ex}{1.3ex plus .2ex}
\titlespacing*{\subsubsection}{0pt}{3ex plus 1ex minus .2ex}{1.3ex plus .2ex}

\title{Tricks from Deep Learning%
\thanks{\textbf{Extended abstract presented at the AD 2016 Conference, Sep 2016, Oxford UK.}}}
\author{Atılım Güneş Baydin\footnote{Corresponding Author, Dept of Computer Science, National
    University of Ireland Maynooth,
    \href{mailto:gunes@cs.nuim.ie}{\texttt{gunes@cs.nuim.ie}}\newline (Current address: Dept of Engineering Science, University of Oxford, \href{mailto:gunes@robots.ox.ac.uk}{\texttt{gunes@robots.ox.ac.uk}})}
  \qquad
  \href{http://barak.pearlmutter.net}{\color{black}Barak A. Pearlmutter}\footnote{Dept of Computer Science, National University of
    Ireland Maynooth,
    \href{mailto:barak@pearlmutter.net}{\texttt{barak@pearlmutter.net}}}
  \qquad
  \href{http://engineering.purdue.edu/~qobi}{\color{black}Jeffrey Mark Siskind}\footnote{School of Electrical and Computer Engineering, Purdue
    University, \href{mailto:qobi@purdue.edu}{\texttt{qobi@purdue.edu}}}}
\date{April 2016}

\begin{document}
\maketitle
\thispagestyle{empty}

\section*{Introduction}

The deep learning \cite{lecun2015deep,Schmidhuber2015,Goodfellow-et-al-2016-Book} community has devised a diverse set of methods to make gradient optimization, using large datasets, of large and highly complex models with deeply cascaded nonlinearities, practical.
Taken as a whole, these methods constitute a breakthrough, allowing computational structures which are quite wide, very deep, and with an enormous number and variety of free parameters to be effectively optimized.
The result now dominates much of practical machine learning, with applications in machine translation, computer vision, and speech recognition.
Many of these methods, viewed through the lens of algorithmic differentiation (AD), can be seen as either addressing issues with the gradient itself, or finding ways of achieving increased efficiency using tricks that are AD-related, but not provided by current AD systems.

The goal of this paper is to explain not just those methods of most relevance to AD, but also the technical constraints and mindset which led to their discovery.
After explaining this context, we present a ``laundry list'' of
methods developed by the deep learning community.
Two of these are discussed in further mathematical detail: a way to dramatically reduce the size of the tape when performing reverse-mode AD on a (theoretically) time-reversible process like an ODE integrator; and a new mathematical insight that allows for the implementation of a stochastic Newton's method.


\subsection*{The Deep Learning Mindset}

The differences in mindset between the AD and deep learning communities are rooted in their different goals and consequent practices. To grossly caricature the situation, the communities have different typical workflows.

\bigskip

\noindent\fbox{\parbox[t]{0.48\textwidth}{
\textbf{A Typical AD Workflow}\medskip\raggedright
\begin{compactenum}
\item Primal computation (e.g., climate simulation) is given.
\item Calculate exact derivatives (e.g., the gradient) automatically and efficiently.
\item Use these derivatives (e.g., for sensitivity analysis or in a standard gradient optimization method).
\item Iterate to improve end-to-end accuracy throughout.
\end{compactenum}
}}
\hfill
\fbox{\parbox[t]{0.48\textwidth}{
\textbf{A Typical Deep Learning Workflow}\medskip\raggedright
\begin{compactenum}
\item Construct primal process whose derivatives are ``nice'', meaning easy to calculate and to use for optimization.
\item Manually code approximate (e.g., stochastic) gradient calculation.
\item Use in custom manually-assisted stochastic gradient optimization method.
\item Iterate to improve generalization on novel data.
\end{compactenum}
}}

\bigskip

\noindent Given these different settings, it is understandable that the deep learning community has developed methods to address two problems.
\begin{enumerate}[(I)]
\item \textbf{Methods for making a forward process whose gradient is ``nice'' in an appropriate sense.}

  The important difference here from the usual situation in AD is that, in the deep learning community, the ``primal'' computational process being trained needs to have two properties: it needs to be sufficiently powerful to perform the desired complex nonlinear computation; and it has to be possible to use training data to set this primal processes' parameters to values that will cause it to perform the desired computation. It is important to note that this parameter setting need not be unique.

  When using gradient methods to set these parameters, there are two dangers.  One is that the system will be poorly conditioned. An intuition for stiffness in this context is that, in a stiff system, changing one parameter requires precise compensatory changes to other parameters to avoid large increases in the optimization criterion. The second danger is that the gradient will be uselessly small for some parameters. Such small gradients would be expected in deeply nested computational process relating inputs to outputs. After all, each stage of processing has a Jacobian. These Jacobians have spectra of singular values. If these singular values are all less than one, then the gradient will be washed out, layer by layer, during the backwards sweep of reverse AD. And if the singular values exceed one, the gradients will instead grow exponentially, leading to a similar sort of catastrophe.

  In response to these difficulties, the deep learning community has come up with a variety of architectural features which are added to the primal process to give it a better behaved gradient, and has also come up with modifications of the reverse AD procedure resulting in the calculation of what we might call a pseudo-gradient, which is designed to be more useful than the true gradient for purposes of optimization.

\item \textbf{Faster and more robust stochastic gradient optimization methods.}

  Simply calculating the actual gradient is so slow in deep learning systems that it is impractical to use the true gradient for purposes of optimization. Instead much faster computations are used to calculate an (unbiased) estimate of the gradient, and these are used for optimization. Such \emph{stochastic optimization} \citep{robbins1951} were first proposed at the dawn of the computer age, but have in practice been something of a black art, requiring constant manual intervention to achieve acceptable performance.  This unfortunate state of affairs has now been rectified \citep{bottou-2010}, with the development of algorithms that are asymptotic improvements over their classical antecedents, as well as methods for automatically controlling the stochastic optimization process not just in its terminal asymptotic quadratic phase (which is in practice of little interest in deep learning systems) but also during the so-called initial transient phase.
\end{enumerate}

\section*{Deep Learning Methods}

We proceed to discuss methods developed by the machine learning community to address each of these issues, in turn.  First, \emph{gradient tricks}, namely methods to make the gradient either easier to calculate or to give it more desirable properties.  And second, \emph{optimization tricks}, namely new methods related to stochastic optimization.

\subsection*{Gradient Tricks}

A na\"ive description of the basic setup in deep learning would be a many-layers-deep network of perceptrons, that is trained by computing the gradient of the training error through reverse mode AD \citep{rumelhart1986learning}, usually with some layers of linear convolutions \citep{Fukushima1980, krizhevsky2012imagenet} and occasionally recurrent structures \citep{pearlmutter1995gradient,Schmidhuber2015}. The reality, however, is not so simple, as such systems would exhibit a number of problems.
\begin{itemize}
\setlength\itemsep{-0.2em}
\item Standard gradient descent with random initial weights would perform poorly with deep neural networks \citep{glorot2010understanding}.
\item Reverse AD blindly applied to the forward model would be highly inefficient with regard to space.
\item Reverse AD, in most implementations, would unroll the matrix operations of the forward model rendering the adjoint calculations quite inefficient.
\item The system would be very poorly conditioned, making standard optimization methods ineffective.
\item Parameters would vary wildly in the magnitude of their gradient, making standard optimization methods ineffective, and perhaps even causing floating point underflow or overflow.
\end{itemize}

Starting with the breakthrough of layer-by-layer unsupervised pretraining (using mainly restricted Boltzmann machines) \citep{hinton2006fast}, the deep learning community has been developing many methods to make gradient descent work with deep architectures. These methods (related to model architectures and gradient calculation) are so numerous, that we cannot hope to even survey the current state of the art in anything shorter than a book.
Here we discuss only a small sample of methods.
Even these benefit from having some structure imposed by dividing them into categories.
These categories are rough, with many methods spanning between a few.

\subsubsection*{Efficiency}
Care is taken to fully utilize the available computational resources, and to choose appropriate tradeoffs between speed, storage, and accuracy.
\begin{enumerate}[(a)]
\item \emph{Commuting $\nabla$ with $\sum$.}
  The gradient operator is linear, and therefore commutes with sum, or equivalently with the average which we will denote with angle brackets $\langle \cdot \rangle$.
In other words, $\nabla_w \langle E( x_p; w) \rangle_p = \langle \nabla_w E( x_p; w) \rangle_p$,
  where $E(x_p;w)$ is the error of the network with parameters $w$ on training pattern $p$.
  If the left hand side of this equation were coded and the gradient calculated with any AD system using reverse AD, the result would be highly inefficient, as the ``forward sweep'' would encompass the entire training set, and would consume enormous storage and would be hard to parallelize.
  The second form, in contrast, lends itself to parallelism across training cases, and each ``little'' gradient requires a constant and very manageable amount of storage.
\item \emph{Stochastic Gradient.}  It is not really feasible to calculate the true gradient when there is a large dataset, as just calculating the true error $E = \langle E_p \rangle = \frac{1}{|D|} \sum_{p\in D} E_p$ over all patterns $p$ in the dataset $D$ is impractically slow.
  Instead a sample is used.
  Calculating the gradient of the error of just a random sample from the dataset (or sequential windows, assuming the dataset has been randomly shuffled) is much faster.
  The expectation of that process is the true value, so this is admissible for use in stochastic gradient optimization.
\item \emph{GPUs, and low-precision numbers.}  Most deep learning work uses GPUs, typically very large clusters of GPUs. 32-, instead of 64-bit, precision floating point is common; and fixed point, with 16-bit or even less precision, is feasible due to the error resiliency of neural networks \citep{gupta2015deep}. 
  No current AD systems can manage the delicate tradeoffs concerning which quantities require more or less precision in their representations.
  Note that derivatives often need more precision, or have a different numeric range, than the primal quantities they are associated with.
\item \emph{Mini-Batches and data parallelism.}  In calculating the stochastic gradient, it is tempting to do the minimal amount of computation necessary to obtain an unbiased estimate, which would involve a single sample from the training set.  In practice it has proven much better to use a block of contiguous samples, on the order of dozens.  So instead of $E_p$ for some $p$, one uses $n^{-1}\sum_{i=0}^{n-1} E_{p+i}$ for some $p$.
  This has two advantages: the first is less noise, and the second is that data-parallelism can be used, with favorable cache properties where every quantity in the network is replaced by an $n$-vector, utilizing SIMD on CPUs and GPUs. Conventional AD systems could not maintain this sort of allocation-free vector parallelism through the reverse AD transform.
\item \emph{Reversible learning.}
  In deep learning it is sometimes desired to perform bi-level optimization, usually in order to tune hyperparameters of the training process.
  Na\"ively, since the primal process is $w(t+1) = w(t) - \eta \nabla_w E(x_t; w(t))$, this would require saving all the $w(1), \ldots, w(t), w(t+1), \ldots, w(T)$.
  Since $w(t)$ may consist of millions of parameters, and $T$ will typically be in the millions, this is not really feasible.
  Fortunately a clever technique has been developed to do reverse AD through this process with less storage \citep{maclaurin2015gradient}.\footnote{Another possibility, not used in the deep learning community, would be checkpoint reverse mode.} The trick is to note that this process looks suspiciously like integration of a time-reversible ODE.  So we could try to run the system backwards while in the reverse AD reverse pass, calculating necessary quantities as needed: $w(t) \approx w(t+1) + \eta \nabla_w E(x_t;w(t+1))$.  The would be approximate due to floating point inaccuracy as well as the use of $w(t+1)$ rather than $w(t)$ in $E(\cdot)$.
  However these inaccuracies would typically be very small, so the difference between these could be computed during the \emph{forward} sweep, and encoded efficiently in a highly compressed form.

  This same technique seems also applicable to performing reverse AD through any process which is (theoretically) time-reversible, such as most ODE or PDE integrators.  Of course, the savings here amount to only a constant factor (albeit perhaps a very large one) over standard reverse AD.  Combining the method with checkpoint reverse would seem natural, and would allow much larger leaf nodes in the checkpoint AD reverse computation graph.
\end{enumerate}

\subsubsection*{Vanishing or Exploding Gradients}
In a multi-layered structure, one would expect the gradients of quantities at early layers to be nearly zero (assuming the gains at intermediate levels are below unity) or to be enormous (assuming the gains are above unity).
Some methods avoid this.
Others work around it, effectively doing a block-structured diagonal pre-conditioning of the gradient.
\begin{enumerate}[(a)]
\item \emph{Rectified linear units instead of sigmoids.}  Classic multi-layer perceptrons use the sigmoid transfer function $\xi \mapsto 1/(1+\exp(-\xi))$, but this has a derivative which goes to zero when $\xi \gg 0$.  That means that when a unit in the network receives a very strong signal, it becomes difficult to change.  Using a rectified linear unit (ReLU) transfer function, $\xi \mapsto \max(0,\xi)$ overcomes this problem, making the system more plastic even when strong signals are present.
Many variants of this have been proposed, often to avoid the lack of continuity in the derivative of the ReLU.  One such are exponential linear units (ELUs) \citep{clevert2015fast}.
\item \emph{Long short-term memory (LSTM).}  A recurrent network, unfolded in time, is simply a deep network with some invariants imposed on the parameter matrices.  One technique that has proven useful in allowing gradient information to span long temporal spans in the context of recurrent networks, or equivalently many layers in a deep network, is the LSTM architecture \citep{hochreiter1997long}.  This is essentially a hold-value unit which can have quantities gated in and out. The recent gated recurrent unit (GRU) model \citep{chung2014empirical} is a simplification of this idea.

\item \emph{Gradient clipping.}  In the domain of deep learning, there are often outliers in the training set: exemplars that are being classified incorrectly, for example, or improper images in a visual classification task, or mislabeled examples, and the like.  These can cause a large gradient inside a single mini-batch, which washes out the more appropriate signals.  For this reason a technique called gradient clipping \citep{pascanu2013difficulty} is often used, in which components of the gradient exceeding a threshold (in absolute value) are pushed down to that threshold.


\end{enumerate}

\subsubsection*{Conditioning}
Keeping the error surface well conditioned for gradient optimization has been one of the keys to the current widespread deployment of deep learning.
\begin{enumerate}[(a)]
\item \emph{Dropout.}  Imagine a network in which multiple units together represent some important feature, requiring a precise weighting of their values in downstream processing.  This would make optimization quite difficult, as it sets up couplings between parameters which must be maintained.  A technique to avoid such ``unfortunate collusions'' is dropout \citep{srivastava2014dropout}, in which each unit in the network is, on each training pattern pass, randomly ``dropped out'' with some probability (typically 50\%) by holding its value at zero.  This encourages detected features to be independently meaningful.  (In ``production mode'' dropout is turned off, and the weights scaled to compensate, to minimize the noise when performance matters.)
\item \emph{Careful initialization.} Considering how the variances of activation values and gradients can be maintained between the layers in a network leads to intelligent normalized initialization schemes, which enable substantially faster optimization convergence \citep{glorot2010understanding}.
\end{enumerate}

\subsection*{Optimization Tricks}
\begin{enumerate}[(a)]
\item \emph{Early stopping.}  When fitting a dynamic system to data, as exact a match as possible is desired, so the true optimum is sought.  This is not the case in machine learning, where the optimization is of error on a training set, while the primary concern is generally not performance on the training set, but on as-yet-unseen new data.  There is often a tradeoff between the two, encountered after optimization has proceeded for a significant amount of time.  This is addressed by early stopping \citep{bengio2012practical}, in which an estimate of performance on unseen data is maintained, and optimization is halted early when this estimated generalization performance stops improving, even if performance on the training set is continuing to improve.  (An estimate of generalization is often obtained by holding back some training data and using it only for this purpose, and never for optimization.)

\item Many AD practitioners have had the demotivating experience of explaining how AD can be used to efficiently calculate Hessian-vector products, only to be asked whether it might be possible to instead calculate the Hessian-inverse-vector product.
  Although an efficient way to perform this calculation exactly remains unknown, a method has been discovered to efficiently calculate an \emph{unbiased estimate} of the Hessian-inverse-vector product \citep{agarwal2016second}!
  Note that we can express a matrix inverse as a series
  \(
      H^{-1} = \sum_{i=0}^{\infty} (I - H)^i
  \)
  where we assume that the spectrum of $H$ allows convergence.
  We could obtain an unbiased estimate of this sum as follows.
  Let $p(i)$ be a distribution with support for all integers $i\geq 0$.
  Choose $i \sim p(i)$.  Calculate $p(i)^{-1}(I-H)^i$.  This is equal, in expectation, to $H^{-1}$.
  Similarly $p(i)^{-1}\overbrace{(I-H)(\cdots((I-H)}^{\text{$i$ times}}v))$ is an unbiased estimate of $H^{-1}v$, and can be computed with $i$ Hessian-vector products.  If $p(\cdot)$ is chosen to have small mean, then $i$ will usually be small.
  If instead of an exact Hessian-vector product we can instead only compute an unbiased estimate of this product, the same procedure will work, except that each $H$ in the above expression becomes $\hat{H}$, an operator that performs a stochastic unbiased Hessian-vector product.
  Further development improves the efficiency of the computation, and embeds it in a stochastic Newton's method with proven efficiency properties.

\end{enumerate}

\section*{Conclusion}

Many advances in AD, both longstanding and recent, are of great potential utility to machine learning in general and deep learning in particular.
Analogously, the machine learning community in general, and the deep learning community in particular, have been using computational derivatives ``in anger'' for quite some time.
They have been forced to build very large systems whose optimization would be intractable using generic AD systems and batch gradient optimization.
Necessity has been the mother of invention, and they have discovered a variety of novel methods which allow them to handle large systems and enormous datasets.
Many of these methods are related to AD in some fashion, and it is our hope that the AD community will find them of interest.

\section*{Acknowledgments}
This work was supported, in part, by Science Foundation Ireland grant
09/IN.1/I2637 and by NSF grant 1522954-IIS.\@
Any opinions, findings, and conclusions or recommendations expressed in this
material are those of the authors and do not necessarily reflect the views
of the sponsors.

\bibliographystyle{unsrtnat}
\begin{small}
  \setlength{\bibsep}{0.2ex}
  \bibliography{ad2016d}
\end{small}











\end{document}